\documentclass[10pt,twocolumn,letterpaper]{article}

\usepackage[review]{cvpr}      

\usepackage{graphicx}
\usepackage{amsmath}
\usepackage{amssymb}
\usepackage{mathtools}
\usepackage{booktabs}
\usepackage{enumitem}
\usepackage{bm}
\usepackage{bbm}
\usepackage{xcolor}
\usepackage{algpseudocode}
\usepackage{algorithm}

\settoggle{cvprfinal}{true}

\usepackage[pagebackref=true,breaklinks=true,letterpaper=true,colorlinks,bookmarks=false ]{hyperref}
\hypersetup{citecolor=[RGB]{119,185,0}}

\usepackage[capitalize]{cleveref}
\crefname{section}{Sec.}{Secs.}
\Crefname{section}{Section}{Sections}
\Crefname{table}{Table}{Tables}
\crefname{table}{Tab.}{Tabs.}

\def\cvprPaperID{00289} 
\def\confName{CVPR}
\def\confYear{2023}
\definecolor{taskcolor}{RGB}{130,183,231}
\definecolor{teachercolor}{RGB}{148,197,113}

\begin{document}

\title{Vision Transformers Are Good Mask Auto-Labelers}

\author{
Shiyi Lan$^{1}$~~\,
Xitong Yang$^{2}$~~\,
Zhiding Yu$^{1}$~~\,
Zuxuan Wu$^{3}$~~\,
Jose M. Alvarez$^{1}$~~\,
Anima Anandkumar$^{1,4}$ \\
$^1$NVIDIA~~\,
$^2$Meta AI, FAIR~~\,
$^3$Fudan University~~\,
$^4$Caltech \\
{
\fontsize{9.4pt}{9.84pt}\selectfont
\url{https://github.com/NVlabs/mask-auto-labeler}
}
}

\makeatletter
\let\@oldmaketitle\@maketitle%
\renewcommand{\@maketitle}{\@oldmaketitle%
\centering
    \vspace{-0.4cm}
    \includegraphics[trim=0 0 0 0,clip,width=1.0\linewidth]{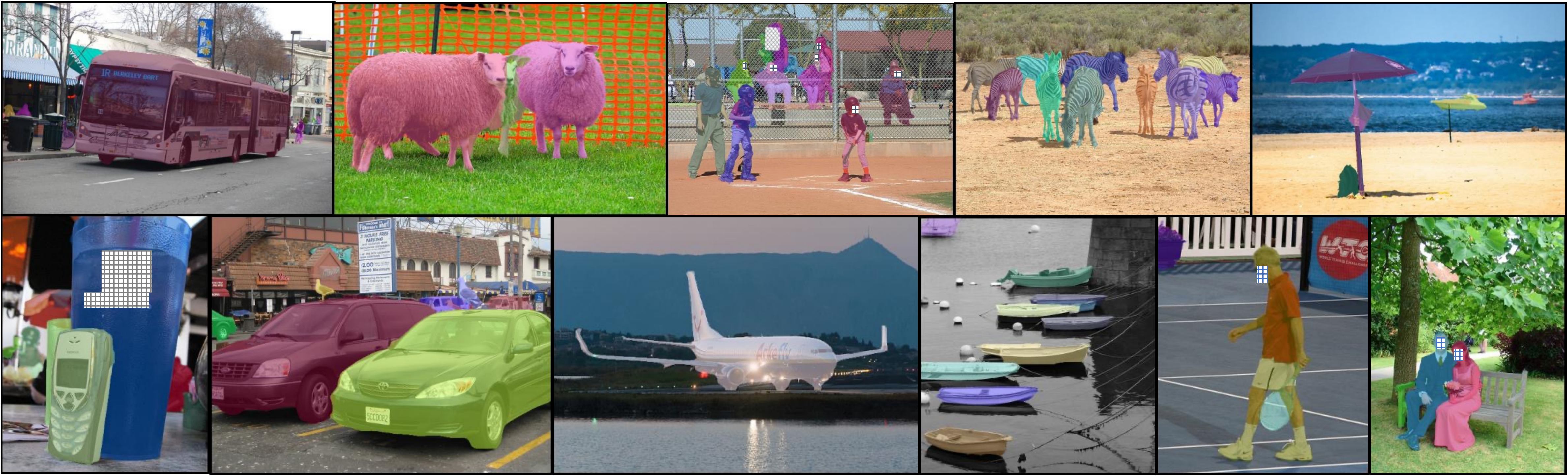}
    \vspace{-0.55cm}
    \captionof{figure}{\textbf{Examples of mask pseudo-labels generated by Mask Auto-Labeler on COCO.} Only human-annotated bounding boxes are used as supervision during training to obtain these results.}
    \label{fig:teaser}
\bigskip}                   %
\makeatother

\maketitle

\begin{abstract}
We propose Mask Auto-Labeler (MAL), a high-quality Transformer-based mask auto-labeling framework for instance segmentation using only box annotations. MAL takes box-cropped images as inputs and conditionally generates their mask pseudo-labels.We show that Vision Transformers are good mask auto-labelers. Our method significantly reduces the gap between auto-labeling and human annotation regarding mask quality. Instance segmentation models trained using the MAL-generated masks can nearly match the performance of their fully-supervised counterparts, retaining up to 97.4\% performance of fully supervised models. The best model achieves 44.1\% mAP on COCO instance segmentation (test-dev 2017), outperforming state-of-the-art box-supervised methods by significant margins. Qualitative results indicate that masks produced by MAL are, in some cases, even better than human annotations.
\end{abstract}


\section{Introduction}
\label{sec:intro}

Computer vision has seen significant progress over the last decade. Tasks such as instance segmentation have made it possible to localize and segment objects with pixel-level accuracy. However, these tasks rely heavily on expansive human mask annotations. For instance, when creating the COCO dataset, about 55k worker hours were spent on masks, which takes about 79\% of the total annotation time~\cite{lin2014microsoft}. Moreover, humans also make mistakes. Human annotations are often misaligned with actual object boundaries. On complicated objects, human annotation quality tends to drop significantly if there is no quality control. Due to the expensive cost and difficulty of quality control, some other large-scale detection datasets such as Open Images~\cite{OpenImages} and Objects365~\cite{shao2019objects365}, only contain partial or even no instance segmentation labels.

In light of these limitations, there is an increasing interest in pursuing box-supervised instance segmentation, where the goal is to predict object masks from bounding box supervision directly. Recent box-supervised instance segmentation methods~\cite{hsu2019weakly,tian2021boxinst,li2022box,lan2021discobox,cheng2022boxteacher} have shown promising performance. The emergence of these methods challenges the long-held belief that mask annotations are needed to train instance segmentation models. However, there is still a non-negligible gap between state-of-the-art approaches and their fully-supervised oracles.

\begin{figure}[t]
\centering
\includegraphics[width=\linewidth]{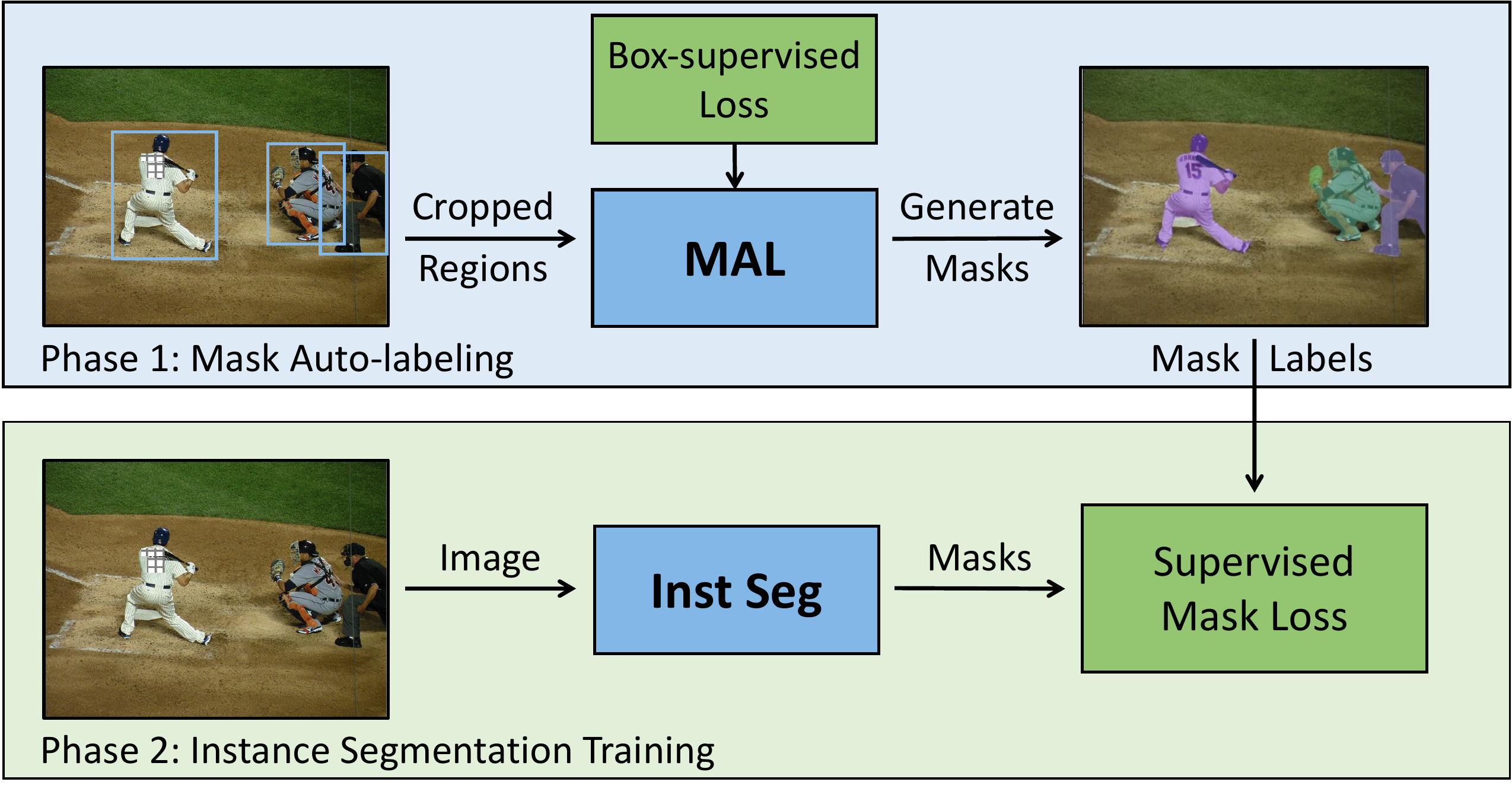}\vspace{-0.1cm}
\caption{An overview of the two-phase framework of box-supervised instance segmentation. For the first phase, we train Mask Auto-Labeler using box supervision and conditionally generate masks of the cropped regions in training images (top). We then train the instance segmentation models using the generated masks (bottom).}
\label{fig:framework}
\vspace{-0.1cm}
\end{figure}

\textbf{Our contributions:} To address box-supervised instance segmentation, we introduce a two-phase framework consisting of a mask auto-labeling phase and an instance segmentation training phase (see Fig.~\ref{fig:framework}). We propose a Transformer-based mask auto-labeling framework, Mask Auto-Labeler (MAL), that takes Region-of-interest (RoI) images as inputs and conditionally generates high-quality masks (demonstrated in Fig.~\ref{fig:teaser}) within the box. Our contributions can be summarized as follows:

\begin{itemize}[leftmargin=1.3em]
\item Our two-phase framework presents a versatile design compatible with any instance segmentation architecture. Unlike existing methods, our framework is simple and agnostic to instance segmentation module designs.

\item We show that Vision Transformers (ViTs) used as image encoders yield surprisingly strong auto-labeling results. 

We also demonstrate that some specific designs in MAL, such as our attention-based decoder, multiple-instance learning with box expansion, and class-agnostic training, crucial for strong auto-labeling performance. Thanks to these components, MAL sometimes even surpasses humans in annotation quality. 

\item Using MAL-generated masks for training, instance segmentation models achieve up to 97.4\% of their fully supervised performance on COCO and LVIS. Our result significantly narrows down the gap between box-supervised and fully supervised approaches. 
We also demonstrate the outstanding open-vocabulary generalization of MAL by labeling novel categories not seen during training.

\end{itemize}

Our method outperforms all the existing state-of-the-art box-supervised instance segmentation methods by large margins.  
This might be attributed to good representations of ViTs and their emerging properties such as meaningful grouping~\cite{caron2021emerging}, where we observe that the attention to objects might benefit our task significantly (demonstrated in Fig. ~\ref{fig:attention}). 
We also hypothesize that our class-agnostic training design enables MAL to
focus on learning general grouping instead of focusing on category
information.
Our strong results pave the way to remove the need for expensive human annotation for instance segmentation in real-world settings.

\section{Related work}

\begin{figure*}[t]
\centering
\includegraphics[width=\linewidth]{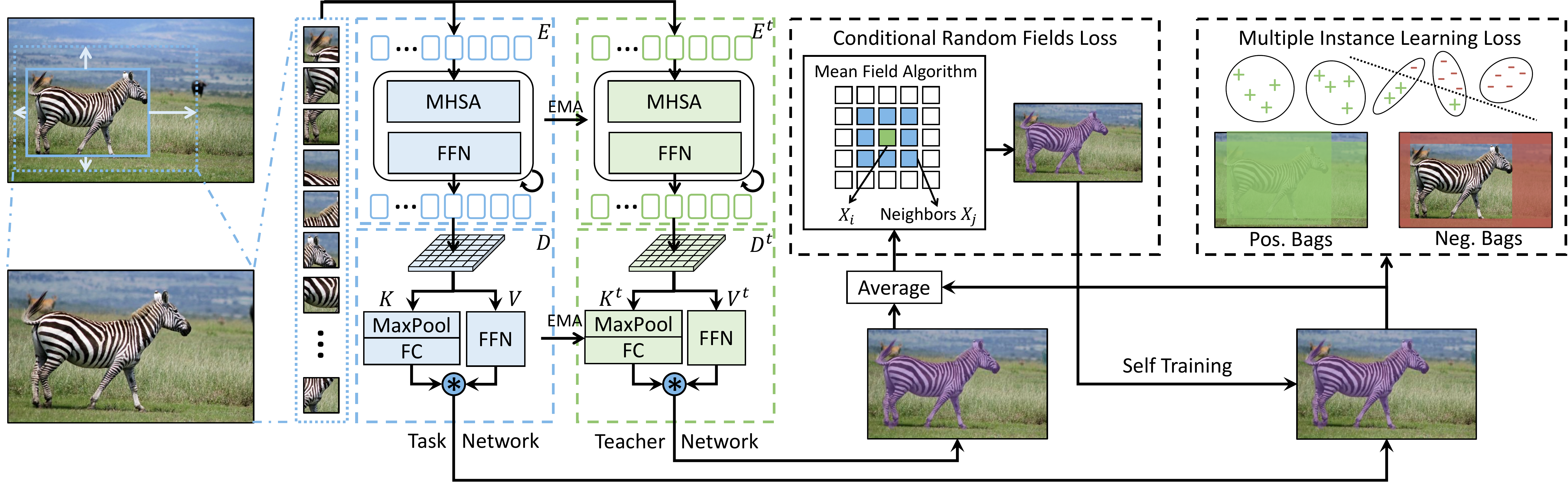}
\caption{Overview of MAL architecture. We visualize the architecture of Mask Auto-Labeler. Mask Auto-Labeler takes cropped images as inputs. Mask Auto-Labeler consists of two symmetric networks, \textcolor{taskcolor}{Task Network} and \textcolor{teachercolor}{Teacher Network}. Each network contains the image encoder $E$(or $E^t$), and the mask decoder $D$(or $D^t$). We use the exponential moving average (EMA) to update the weights of the teacher network. We apply multiple instance learning (MIL) loss and conditional random fields (CRFs) loss. The CRF loss takes the average mask predictions of the teacher network and the task network to make the training more stable and generate refined masks for self-training. }
\label{fig:mal}
\vspace{-1em}
\end{figure*}

\subsection{Vision Transformers}

Transformers were initially proposed in natural language processing~\cite{vaswani2017attention}. Vision Transformers~\cite{dosovitskiy2020image} (ViTs) later emerged as highly competitive visual recognition models that use multi-head self-attention (MHSA) instead of convolutions as the basic building block. These models are recently marked by their competitive performance in many visual recognition tasks~\cite{liu2021swin}. We broadly categorize existing ViTs into two classes: plain ViTs, and hierarchical ViTs.

\vspace{0.1cm}
\noindent\textbf{Standard Vision Transformers.} Standard ViTs~\cite{dosovitskiy2020image} are the first vision transformers. Standard ViTs have the simplest structures, which consist of a tokenization embedding layer followed by a sequence of MHSA layers. However, global MHSA layers can be heavy and usually face significant optimization issues. To improve their performance, many designs and training recipes are proposed to train ViTs in data-efficient manners~\cite{caron2021emerging, he2022masked, bao2021beit, wang2022image, touvron2021training, Touvron2022DeiTIR, Touvron_2021_ICCV, meng2022adavit}.

\vspace{0.1cm}
\noindent\textbf{Hierarchical Vision Transformers.} 
Hierarchical Vision Transformers~\cite{wang2021pyramid,wang2022pvt,liu2021swin,liu2022swin} are pyramid-shaped architectures that aim to benefit other tasks besides image classification with their multi-scale designs. On top of plain ViTs, these ViTs~\cite{wang2021pyramid,wang2022pvt} separate their multi-head self-attention layers into hierarchical stages. Between the stages, there are spatial reduction layers, such as max-pooling layers. These architectures are usually mixed with convolutional layers~\cite{zhou2021deepvit} and often adopt efficient self-attention designs to deal with long sequence lengths.

\subsection{Instance segmentation}

Instance segmentation is a visual recognition task that predicts the bounding boxes and masks of objects.

\vspace{0.1cm}
\noindent\textbf{Fully supervised instance segmentation.} In this setting, both bounding boxes and instance-level masks are provided as the supervision signals. Early works~\cite{he2017mask, li2017fully,xu2019deepmask,hu2017fastmask} follow a two-stage architecture that generates box proposals or segmentation proposals in the first stage and then produces the final segmentation and classification information in the second stage. Later, instance segmentation models are broadly divided into two categories: some continue the spirit of the two-stage design and extend it to multi-stage architectures~\cite{cai2018cascade, chen2019hybrid}. Others simplify the architecture and propose one-stage instance segmentation, e.g., YOLACT~\cite{bolya2019yolact}, SOLO~\cite{wang2020solov2,wang2021solo}, CondInst~\cite{tian2020conditional}, PolarMask~\cite{xie2020polarmask,xie2021polarmask++}. Recently, DETR and Deformable DETR~\cite{carion2020end,zhu2021deformable} show great potential of query-based approaches in object detection. Then, methods like MaxDeepLab~\cite{wang2021max}, MaskFormer~\cite{cheng2021per}, PanopticSegFormer~\cite{li2022panoptic}, Mask2Former~\cite{cheng2021mask2former} and Mask DINO~\cite{li2022mask} are introduced along this line and have pushed the boundary of instance segmentation. On the other hand, the instance segmentation also benefits from more powerful backbone designs, such as Swin Transformers~\cite{liu2021swin,liu2022swin}, ViTDet~\cite{li2022exploring}, and ConvNeXt~\cite{liu2022convnet}.

\vspace{0.1cm}
\noindent\textbf{Weakly supervised instance segmentation.} There are two main styles of weakly supervised instance segmentation: learning with image-level and box-level labels. The former uses image-level class information to perform instance segmentation~\cite{durand2017wildcat,jin2017webly,ahn2018learning,fan2018associating,sun2020mining}, while the latter uses box-supervision. Hsu et al.~\cite{hsu2019weakly} leverages the tight-box priors. Later, BoxInst~\cite{tian2021boxinst} proposes to leverage color smoothness to improve accuracy. Besides that, DiscoBox~\cite{lan2021discobox} proposes to leverage both color smoothness and inter-image correspondence for the task. Other follow-ups~\cite{cheng2022boxteacher, li2022box} also leverage tight-box priors and color smoothness priors.

\vspace{0.1cm}
\subsection{Deep learning interpretation}
The interest in a deeper understanding of deep networks has inspired many works to study the interpretation of deep neural networks. For example, Class Activation Map (CAM)~\cite{zhou2016learning} and Grad-CAM~\cite{selvaraju2017grad} visualize the emerging localization during image classification training of convolutional neural networks (CNNs). This ability has also inspired much weakly-supervised localization and shows deep connections to general weakly-supervised learning, which partly motivates our decoder design in this paper. DINO~\cite{caron2021emerging} further shows that meaning visual grouping emerges during self-supervised learning with ViTs. In addition, FAN~\cite{zhou2022understanding} shows that such emerging properties in ViTs are linked to their robustness.

\section{Method}

Our work differs from previous box-supervised instance segmentation frameworks~\cite{hsu2019weakly,tian2021boxinst,lan2021discobox,li2022box,cheng2022boxteacher} that simultaneously learns detection and instance segmentation. We leverage a two-phase framework as visualized in Fig.~\ref{fig:framework}, which allows us to have a network focused on generating mask pseudo-labels in phase 1, and another network focused on learning instance segmentation~\cite{he2017mask,cai2018cascade,li2022exploring,cheng2021mask2former} in phase 2. Our proposed auto-labeling framework is used in phase 1 to generate high-quality mask pseudo-labels.

We propose this two-phase framework because it brings the following benefits:
\begin{itemize}[leftmargin=1.3em]
    \item We can relax the learning constraints in phase 1 and focus only on mask pseudo-labels. Therefore, in this phase, we can take Region-of-interest (RoI) images instead of untrimmed images as inputs. This change allows us to use a higher resolution for small objects and a strong training technique mentioned in Sec.~\ref{sec3:input_preprocessing}, which helps improve the mask quality.
    \item We can leverage different image encoders and mask decoders in phases 1 and 2 to achieve higher performance. We empirically found that phases 1 and 2 favor different architectures for the image encoders and mask decoders. See the ablation study in Tab.~\ref{tab:mask_decoder} and~\ref{tab:encoder_variation}.
    \item We can use MAL-generated masks to directly train the most fully supervised instance segmentation models in phase 2. This makes our approach more flexible than previous architecture-specific box-supervised instance segmentation approaches~\cite{hsu2019weakly, tian2021boxinst,lan2021discobox,li2022box,cheng2022boxteacher}.
\end{itemize}

As phase 2 follows the previous standard pipelines, which do not need to be re-introduced here, we focus on introducing phase 1 (MAL) in the following subsections.

\subsection{RoI input generation}\label{sec3:input_preprocessing}
Most box-supervised instance segmentation approaches~\cite{hsu2019weakly,tian2021boxinst,lan2021discobox,li2022box} are trained using the entire images. However, we find that using RoI images might have more benefits in box-supervised instance segmentation. Moreover, we compare two intuitive sampling strategies of RoI images to obtain foreground and background pixels and explain the better strategy, box expansion, in detail.

\vspace{0.1cm}
\noindent{\textbf{Benefits of using RoI inputs.}} There are two advantages of using RoI images for inputs. First, using the RoI images as inputs is naturally good for handling small objects because no matter how small the objects are, the RoI images are enlarged to avoid the issues caused by low resolution. Secondly, having RoI inputs allows MAL to focus on learning segmentation and avoid being distracted from learning other complicated tasks, e.g., object detection.
\vspace{0.1cm}
\noindent{\textbf{RoI sampling strategy.}} 
The sampling strategy should ensure both positive and negative pixels are included. We present two straightforward sampling strategies:

\begin{itemize}[leftmargin=1.3em]
    \item The first strategy is to use bounding boxes to crop the images for positive inputs. We crop the images using randomly generated boxes containing only background pixels for negative inputs.
    MAL does not generate good mask pseudo-labels with cropping strategy. We observe that the networks tend to learn the trivial solution (all pixels are predicted as either foreground or background).
    \vspace{-1em}
    \item The second is to expand the bounding boxes randomly and include background pixels, where negative bags are chosen from the expanded rows and columns. We visualize how we define positive/negative bags in Fig.~\ref{fig:mal} and explain the detail in Sec.~\ref{sec:losses}. \textbf{This detailed design is critical to make MAL work} as it prevents MAL from learning trivial solutions. Without this design, the generated masks tend to fill the entire bounding box. 
\end{itemize}

\noindent\textbf{Box expansion specifics.} 
 Given an untrimmed image $\bm{I}^u \in \mathbb{R}^{C \times H^u \times W^u}$ and the bounding box $\bm{b} = ( x_0, y_0, x_1, y_1 )$ indicating the x, y coordinates of the top-left corners and the bottom-right corners. To obtain background pixels, we randomly expand the bounding box $\bm{b}$ to $\bm{b^\prime} = ( x_c + \beta_x (x_0 - x_c), y_c + \beta_x^\prime (y_0 - y_c), x_c + \beta_y (x_1 - x_c), y_c + \beta_y^\prime (y_1 - y_c) )$, where $x_c = (x_0 + x_1) / 2$, $y_c = (y_0 + y_1)/2$. To generate random values of $\beta_x,\beta_x^\prime, \beta_y, \beta_y^\prime$,  we randomly generate
$\theta_x, \theta_y \in [ 0, \theta ]$ for x- and y-direction, where $\theta$ is the upper bound of box expansion rate. Next, we randomly generate $\beta_x \in [0, \theta_x]$ and $\beta_y \in [0, \theta_y]$. In the end, we assign $\beta_x^\prime$ as $\theta_x - \beta_x$ and $\beta_y^\prime$ as $\theta_y - \beta_y$. Finally, we use $\bm{b}^\prime$ to crop the image and obtain trimmed image $\bm{I}^t$. We conduct the ablation study for $\theta$ in Tab.~\ref{tab:box_expansion}. At last, We resize the trimmed image $\bm{I}^t$ to the size of $C \times H^c \times W^c$ as the input image $\bm{I}^c$.

\vspace{1em}

\subsection{MAL architecture}~\label{sec:mal}
\vspace{-1.5em}

\begin{figure}[t]
\centering
\includegraphics[width=\linewidth]{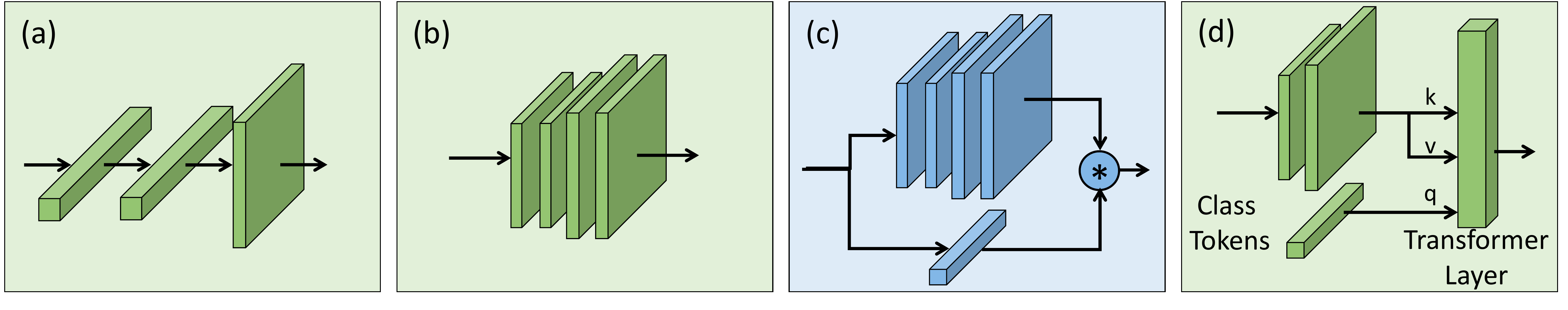}
\caption{(a) The fully connected decoder (b) The fully convolutional Decoder (c) The attention-based decoder (used in MAL) (d) The query-based Decoder.}
\label{fig:mask_decoder}
\vspace{-2em}
\end{figure}

MAL can be divided into two symmetric networks: the task network and the teacher network. The task network consists of an image encoder denoted as $E$, and a mask decoder denoted as $D$, demonstrated in Fig.~\ref{fig:mal}. The architecture of the teacher network is identical to the task network. We denote the segmentation output of the task network and the teacher network as $\bm{m},\bm{m}^t \in \{0,1\}^{N}$, respectively.

\noindent{\textbf{Image encoder}.} We use Standard ViTs~\cite{dosovitskiy2020image} as the image encoder and drop the classification head of Standard ViTs. We compare different image encoders in Sec.~\ref{sec:encoder_variation}. We also try feature pyramid networks on top of Standard ViTs, e.g., FPN~\cite{lin2017feature}, but it causes a performance drop. Similar conclusions were also found in ViTDet~\cite{li2022exploring}.

\noindent\textbf{Mask decoder.}  For the mask decoder $D$, we use a simple attention-based network inspired by YOLACT~\cite{bolya2019yolact}, which includes an instance-aware head $K$ and a pixel-wise head $V$, 
where 
$D(E(\bm{I})) = K(E(\bm{I})) \cdot V(E(\bm{I}))$, 
and `` $\cdot$ '' 
represents the inner-product operator.

For the instance-aware head $K$, we use a max-pooling layer followed by a fully connected layer. The input channel dimension of $K$ is equivalent to the output channel dimension of $E$. The output channel dimension of $K$ is 256.

For the pixel-wise head $V$, we use four sequential convolutional layers. Each is followed by a ReLU layer. Between the second and the third convolutional layer, we insert a bi-linear interpolation layer to increase the feature resolution by 2. The input channel dimension is equivalent to the output channel dimension of $E$. We use 256 dimensions for hidden channels and output channels. We also compare different design choices of mask decoders in Sec.~\ref{sec:mask_decoder}.

\noindent\textbf{Exponential moving average (EMA) teacher.}  Instead of training the teacher network directly, we leverage exponential moving averages (EMA) to update the parameters in the teacher network using the parameters in the task network similar to MOCO~\cite{He_2020_CVPR}. The goal of using EMA Teacher is to eliminate the loss-explosion issues in training since optimizing Standard Vision Transformers is usually non-trivial~\cite{touvron2021training,bao2021beit,he2022masked}. We do not observe any significant performance drop or improvement on DeiT-small-based MAL after removing the teacher network. However, it makes the training more stable when we use larger-scale image encoders in MAL, e.g. ViT-MAE-Base~\cite{he2022masked}.

\subsection{Losses}~\label{sec:losses}
We use Multiple Instance Learning Loss $\mathcal{L}_\text{mil}$ and Conditional Random Field Loss $\mathcal{L}_\text{crf}$ as the box-supervised loss:
\vspace{-0.2cm}
\begin{equation}
    \mathcal{L} = \alpha_\text{mil} \mathcal{L}_\text{mil} + \alpha_\text{crf} \mathcal{L}_\text{crf}
\end{equation}
\vspace{-0.2cm}

\noindent{\textbf{Multiple Instance Learning Loss.}}
The motivation of the Multiple Instance Segmentation is to exploit the priors of tight-bounding box annotations.

After the student network produces the output $\bm{m}$, we apply the Multiple Instance Learning (MIL) Loss on the output mask $\bm{m}$. We demonstrate the process in Fig.~\ref{fig:mal}.

We denote $\bm{m}_{i,j}$ as the mask score at the location $i, j$ in the image $\bm{I}^c$. We define each pixel as an instance in the MIL loss. Inspired by BBTP~\cite{hsu2019weakly}, we treat each row or column of pixels as a bag. We determine whether a bag is positive or negative based on whether it passes a ground-truth box. We define the bags as $\bm{B}$, and each bag $\bm{B}_i$ contains a row or column of pixels. Additionally, we define the label for each bag $\bm{g}$, and each label $\bm{g}_i$ corresponds to a bag $\bm{B}_i$.

Therefore, we use the max pooling as the reduction function and dice loss~\cite{sudre2017generalised}:
\vspace{-0.1cm}
\begin{small}
\begin{equation}
    \mathcal{L}_{mil} = 1- \frac{2 \sum_i \bm{g}_i \cdot \max \{ \bm{B}_i \}^2 }{\sum_i \max\{ \bm{B}_i \}^2 + \sum_i \bm{g}_i^2}
\end{equation}
\end{small}
\vspace{-0.1cm}

\noindent{\textbf{Conditional Random Field Loss.}} The goal of CRF loss is to refine the mask prediction by imposing the smoothness priors via energy minimization. Then, we leverage this refined mask as pseudo-labels to self-train the mask prediction in an online-teacher manner. We use the average mask prediction $\bm{m}^a = \frac{1}{2}(\bm{m} + \bm{m}^t)$ as the mask prediction to be refined for more stable training.

Next, we define a random field $\bm{X} = \{\bm{X}_1, ..., \bm{X}_N\}$, where $N = H^c \times W^c$ is the size of cropped image and each $\bm{X}_i$ represents the label that corresponds to a pixel in $\bm{I}^c$, therefore we have $\bm{X} \in \{ 0, 1\}^N$, meaning the background or the foreground. We use $\bm{l} \in \{ 0, 1\}^{N}$ to represent a labeling of $\bm{X}$ minimizing the following CRF energy:

\vspace{-0.2cm}
\begin{small}
\begin{equation}
    E(\bm{l} | \bm{m}^a, \bm{\bm{X}^c}) = \mu(\bm{X|\bm{m}^a, \bm{I}^c}) + \psi(\bm{X}| \bm{I}^c),
\end{equation}
\end{small}
\vspace{-0.2cm}

\noindent where $\mu(\bm{X|\bm{m}^a, \bm{I}^c})$ represents the unary potentials, which is used to align $\bm{X}_i$ and $\bm{m}^a_i$ since we assume that most of the mask predictions are correct. Meanwhile, $\psi(\bm{X}| \bm{I}^c)$ represents the pairwise potential, which sharpens the refined mask. Specifically, we define the pairwise potentials as:

\vspace{-0.35cm}
\begin{small}
\begin{equation}
    \psi(\bm{X}| \bm{I}^c) = \sum_{\substack{i \in \{0 .. N-1\},\\ j \in \mathcal{N}(i)}} \omega \exp(\frac{-{| \bm{I}^c_i - \bm{I}^c_j|^2}}{2\zeta^2})[\bm{X}_i \neq \bm{X}_j],
\end{equation}
\end{small}
\vspace{-0.25cm}

\noindent where $\mathcal{N}(i)$ represents the set of 8 immediate neighbors to $\bm{X}_i$ as shown in Fig.~\ref{fig:mal}. Then, we use the MeanField algorithm~\cite{krahenbuhl2011efficient,lan2021discobox} to efficiently approximate the optimal solution, denoted as $\bm{l} = MeanField(\bm{I^c},  \bm{m^a})$. We attach the derivation and PyTorch code in the supplementary.  At last, we apply Dice Loss to leverage the refined masks $\bm{l}$ to self-train the models as:
\vspace{-0.15cm}
\begin{small}
\begin{equation}
    \mathcal{L}_\textit{crf} = 1 - \frac{2 \sum_i \bm{l}_i \bm{m}_i}{\sum_i \bm{l}_i^2 + \bm{m}_i^2}
\end{equation}
\end{small}
\vspace{-0.25cm} 

\section{Experiments}

\begin{table*}[t]
\centering
\resizebox{\textwidth}{!}{
\addtolength{\tabcolsep}{3pt}
\begin{tabular}{lccccccccc}
\toprule
Method & Labeler Backbone & InstSeg Backbone & InstSeg Model & ~~~~Sup~~~~ & $(\%)\text{Mask AP}_\text{val}$ & $(\%)\text{Mask AP}_\text{test}$ & $~~(\%)\text{Ret.}_\text{val}~~$ & $~~(\%)\text{Ret.}_\text{test}~~$ \\
\midrule

Mask R-CNN$^*$~\cite{he2017mask} & - & ResNet-101 & Mask R-CNN & Mask & 38.6 & 38.8 & - & - \\
Mask R-CNN$^*$~\cite{he2017mask} &  - & ResNeXt-101 & Mask R-CNN & Mask & 39.5 & 39.9 & - & - \\

CondInst~\cite{tian2020conditional} & - & ResNet-101 & CondInst & Mask  & 38.6 & 39.1 & - & - \\
SOLOv2~\cite{wang2020solov2} & - & ResNet-50 & SOLOv2 & Mask & 37.5 & 38.4 & - & - \\
SOLOv2~\cite{wang2020solov2} & - & ResNet-101-DCN & SOLOv2 & Mask &  41.7 & 41.8  & - & - \\
SOLOv2~\cite{wang2020solov2} & - & ResNeXt-101-DCN & SOLOv2 & Mask &  42.4 & 42.7 & - & - \\
ConvNeXt~\cite{liu2022convnet} & - & ConvNeXt-Small~\cite{liu2022convnet} & Cascade R-CNN & Mask & 44.8 & 45.5 & - & - \\
ConvNeXt~\cite{liu2022convnet} & - & ConvNeXt-Base~\cite{liu2022convnet} & Cascade R-CNN & Mask & 45.4 & 46.1 & - & - \\
Mask2Former~\cite{cheng2021mask2former} & - & Swin-Small & Mask2Former & Mask & \textbf{46.1} & \textbf{47.0} & - & - \\
\midrule
BBTP$\dagger$~\cite{hsu2019weakly} & - & ResNet-101 & Mask R-CNN  & Box & - & 21.1 & - & 59.1 \\
BoxInst~\cite{tian2021boxinst} & - & ResNet-101 & CondInst & Box & 33.0 & 33.2 & 85.5 & 84.9 \\
BoxLevelSet~\cite{li2022box} & - & ResNet-101-DCN & SOLOv2 & Box & 35.0 & 35.4 & 83.9 & 83.5\\
DiscoBox~\cite{lan2021discobox} & - & ResNet-50 & SOLOv2 & Box & 30.7 & 32.0 & 81.9 & 83.3 \\
DiscoBox~\cite{lan2021discobox} & - & ResNet-101-DCN & SOLOv2 & Box & 35.3 & 35.8 & 84.7 & 85.9\\
DiscoBox~\cite{lan2021discobox} & - & ResNeXt-101-DCN & SOLOv2 & Box & 37.3 & 37.9 & 88.0 & 88.8 \\
BoxTeacher~\cite{cheng2022boxteacher} & - & Swin-Base & CondInst & Box & - & 40.0 & - & - \\
\midrule
Mask Auto-Labeler & ViT-MAE-Base~\cite{he2022masked} & ResNet-50 & SOLOv2 & Box & 35.0 & 35.7 & 93.3 & 93.0 \\
Mask Auto-Labeler & ViT-MAE-Base~\cite{he2022masked} & ResNet-101-DCN & SOLOv2 & Box & 38.2 & 38.7 & 91.6 & 92.6\\
Mask Auto-Labeler & ViT-MAE-Base~\cite{he2022masked} & ResNeXt-101-DCN & SOLOv2 & Box & 38.9 & 39.1 & 91.7 & 91.6\\
Mask Auto-Labeler & ViT-MAE-Base~\cite{he2022masked} & ConvNeXt-Small~\cite{liu2022convnet} & Cascade R-CNN & Box & 42.3 & 43.0 & 94.4 & \textbf{94.5} \\
Mask Auto-Labeler & ViT-MAE-Base~\cite{he2022masked} & ConvNeXt-Base~\cite{liu2022convnet} & Cascade R-CNN & Box & 42.9 & 43.3 & \textbf{94.5} & 93.9 \\
Mask Auto-Labeler & ViT-MAE-Base~\cite{he2022masked} & Swin-Small~\cite{liu2021swin} & Mask2Former~\cite{cheng2021mask2former} & Box & \textbf{43.3} & \textbf{44.1} & 93.9 & 93.8 \\
\bottomrule
\end{tabular}
}
\vspace{-0.5em}
\caption{
Main results on COCO. Ret means the retention rate of $\frac{\text{box-supervised mask AP}}{\text{supervised mask AP}}$. MAL with SOLOv2/ResNeXt-101 outperforms DiscoBox with SOLOv2/ResNeXt-101 by 1.6\% on val2017 and 1.3\% on test-dev. Our best model (Mask2former/Swin-Small) achieves 43.3\% AP on val and 44.1\% AP on test-dev.  
}
\label{tab:coco_exp}
\vspace{-1em}
\end{table*}

We evaluate MAL on COCO dataset~\cite{lin2014microsoft}, and LVIS~\cite{gupta2019lvis}. The main results on COCO and LVIS are shown in Tab.~\ref{tab:coco_exp} and ~\ref{tab:lvis_exp}. The qualitative results are shown in Fig.~\ref{fig:teaser} and Fig.~\ref{fig:lvis}.

\subsection{Datasets}

\noindent{\textbf{COCO dataset.}} contains 80 semantic categories. We follow the standard partition, which includes train2017 (115K images), val2017 (5K images), and test-dev (20k images).

\noindent{\textbf{LVIS dataset.}} contains 1200+ categories and 164K images. We follow the standard partition of training and validation.

\subsection{Implementation Details}

We use 8 NVIDIA Tesla V100s to run the experiments.

\noindent{\textbf{Phase 1 (mask auto-labeling).}} We use AdamW~\cite{loshchilov2017decoupled} as the network optimizer and set the two momentums as 0.9, 0.9. We use the cosine and annealing scheduler to adjust the learning rate, which is set to $1.5 \cdot 10^{-6}$ per image. The MIL loss weight $\alpha_\text{mil}$, CRF loss weight $\alpha_\text{crf}$, $\zeta$, and $\omega$ in CRF pairwise potentials are set to 4, 0.5, 0.5, 2, respectively. We analyze the sensitivity of the loss weights and CRF hyper-parameters in Fig. ~\ref{fig:sensitivity}. We use the input resolution of $512 \times 512$, and a batch size of 32 (4 per GPU). For EMA, we use a momentum of 0.996. For the task and teacher network, we apply random flip data augmentation. On top of that, we apply extra random color jittering, random grey-scale conversion, and random Gaussian blur for the task network. We train MAL for 10 epochs. It takes around 23 hours and 35 hours to train MAL with Standard ViT-Base~\cite{dosovitskiy2020image} on the COCO and LVIS datasets, respectively. 

\noindent{\textbf{Phase 2 (Training instance segmentation models).}} We select a couple of high-performance fully supervised instance segmentation models, which are ConvNeXts~\cite{liu2022convnet} with Cascade R-CNN~\cite{cai2018cascade}, Swin Transformers~\cite{liu2021swin} with Mask2Former~\cite{cheng2021mask2former}, ResNets~\cite{he2016deep} and ResNeXts~\cite{xie2017aggregated} with SOLOv2~\cite{wang2020solov2}. MAL works extremely well with these architectures, which demonstrates the great power of Mask Auto-Labeler from the perspective of accuracy and generalization. We leverage the codebase in MMDetection~\cite{mmdetection} for phase 2. Again, we only replace the GT masks with MAL-generated mask pseudo-labels to adjust all these fully supervised models to box-supervised learning.

\subsection{Instance segmentation results}

\noindent{\textbf{Retention Rate.} We argue that the sole mAP of instance segmentation is not fair enough to evaluate box-supervised instance segmentation since the performance gain can be achieved by improving box quality unrelated to segmentation quality. However, the retention rate can better reflect the real mask quality because the fully supervised counterparts also get boosted by the better box results.

\noindent{\textbf{Results on COCO.}} In table~\ref{tab:coco_exp}, we show that various modern instance segmentation models can achieve up to 94.5\% performance with the pseudo-labels of the fully supervised oracles. Our best results are 43.3\% mAP on COCO test-dev and 44.1\% mAP on COCO val, achieved by using MAL (Standard ViT-Base~\cite{dosovitskiy2020image} pretrained with MAE) for phase 1, and using Mask2Former (Swin-Small)~\cite{liu2021swin,cheng2021mask2former} for phase 2.
There is no significant retention drop when we use the mask pseudo-labels to train more powerful instance segmentation models. On the contrary, the higher retention rates on COCO are achieved by the heavier instance segmentation models, e.g., Cascade R-CNN with ConvNeXts and Mask2Former with Swin-Small.  However, other methods have significantly lower retention rates compared with MAL. The experiment results quantitatively imply that the mask quality outperforms other methods by a large margin. 

\noindent{\textbf{Results on LVIS.}} In table~\ref{tab:lvis_exp}, we also observe that all instance segmentation models work very well with the mask pseudo-labels generated by MAL (Ret. = 93\% \textasciitilde \  98\%).  We visualize part of the results in figure ~\ref{fig:lvis}. We also evaluate the open-vocabulary ability of MAL by training MAL on COCO dataset but generating mask pseudo-labels on LVIS, and thus training instance segmentation models using these mask pseudo-labels.

\begin{figure*}[t]
\centering
\includegraphics[width=1.0\linewidth]{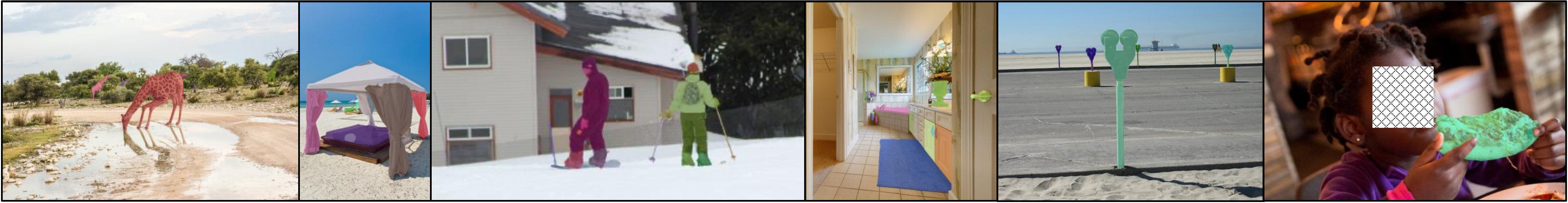}
\vspace{-2em}
\caption{Qualitative results of mask pseudo-labels generated by Mask Auto-Labeler on LVIS v1.}
\label{fig:lvis}
\vspace{-0.5em}
\end{figure*}

\begin{table*}[t]
\centering
\resizebox{\textwidth}{!}{
\addtolength{\tabcolsep}{7pt}
\begin{tabular}{lcccccccc}
\toprule
Method & Autolabeler Backbone & InstSeg Backbone & InstSeg Model & Training Data & Sup & $(\%)\text{Mask AP}_\text{val}$ & $(\%)\text{Ret.}_\text{val}$ \\
\midrule
Mask R-CNN~\cite{he2017mask} & - & ResNet-50-DCN & Mask R-CNN~\cite{he2017mask} & - & Mask & 21.7 & - \\
Mask R-CNN~\cite{he2017mask} & - & ResNet-101-DCN & Mask R-CNN~\cite{he2017mask} & - & Mask & 23.6 & - \\
Mask R-CNN~\cite{he2017mask} & - & ResNeXt-101-32x4d-FPN & Mask R-CNN~\cite{he2017mask} & - & Mask & 25.5 & - \\
Mask R-CNN~\cite{he2017mask} & - & ResNeXt-101-64x4d-FPN & Mask R-CNN~\cite{he2017mask} & - & Mask & 25.8 & - \\
\midrule
Mask Auto-Labeler & ViT-MAE-Base~\cite{he2022masked} & ResNet-50-DCN & Mask R-CNN~\cite{he2017mask} & LVIS v1 & Box & 20.7 & 95.4 \\
Mask Auto-Labeler & ViT-MAE-Base~\cite{he2022masked} & ResNet-101-DCN & Mask R-CNN~\cite{he2017mask} & LVIS v1 & Box & 23.0 & \textbf{97.4} \\
Mask Auto-Labeler  & ViT-MAE-Base~\cite{he2022masked} & ResNeXt-101-32x4d-FPN & Mask R-CNN~\cite{he2017mask} & LVIS v1 & Box & 23.7 & 92.9 \\
Mask Auto-Labeler  & ViT-MAE-Base~\cite{he2022masked} & ResNeXt-101-64x4d-FPN & Mask R-CNN~\cite{he2017mask} & LVIS v1 & Box & \textbf{24.5} & 95.0 \\
\midrule
Mask Auto-Labeler  & ViT-MAE-Base~\cite{he2022masked} & ResNeXt-101-32x4d-FPN & Mask R-CNN~\cite{he2017mask} & COCO & Box & 23.3 & 91.8 \\
Mask Auto-Labeler & ViT-MAE-Base~\cite{he2022masked} & ResNeXt-101-64x4d-FPN & Mask R-CNN~\cite{he2017mask} & COCO & Box & \textbf{24.2} & \textbf{93.8} \\
\bottomrule
\end{tabular}
}
\vspace{-1em}
\caption{Main results on LVIS v1. Training data means the dataset we use for training MAL.  We also finetune it on COCO and then generate pseudo-labels of LVIS v1. Compared with trained on LVIS v1 directly, MAL finetuned on COCO only caused around 0.35\% mAP drop on the final results, which indicates the great potential of the open-set ability of MAL.  Ret means the retention rate of $\frac{\text{box-supervised mask AP}}{\text{supervised mask AP}}$. 
}
\label{tab:lvis_exp}
\vspace{-0.5em}
\end{table*}

\subsection{Image encoder variation}\label{sec:encoder_variation}
To support our claim that Vision Transformers are good auto-labelers, we compare three popular networks as the image encoders of MAL: Standard Vision Transformers~\cite{dosovitskiy2020image, touvron2021training, he2022masked}, Swin Transformer~\cite{liu2021swin}, ConvNeXts~\cite{liu2022convnet} in Tab.~\ref{tab:encoder_variation}. 

First, we compare the fully supervised pretrained weights of these three models. We choose the official fully supervised pre-trained weights of ConvNeXts and Swin Transformers. For Standard Vision Transformers, we adopt a popular fully supervised approach, DeiT~\cite{touvron2021training}. We observe that fully supervised Standard Vision Transformers (DeiT) as image encoders of Mask Auto-Labeler are better than Swin Transformers and ConvNeXts even though the imaganet-1k performance of Swin Transformers and ConvNeXts is higher than that of DeiT. We argue that the success of Standard Vision Transformers might be owed to the self-emerging properties of Standard ViTs~\cite{dosovitskiy2020image, caron2021emerging} (visualized in Fig.~\ref{fig:attention}), and the larger-receptive field brought by global multi-head self-attention layers.

Second, the mask pseudo-labels can be further improved by Mask AutoEncoder (MAE) pretraining~\cite{he2022masked}. The potential reason might be that MAE pretraining enhances Standard ViTs via learning pixel-level information, which is very important for dense-prediction tasks like segmentation.

\subsection{Mask decoder variation}\label{sec:mask_decoder}

\begin{figure*}[t]
\centering
\includegraphics[width=1.0\linewidth]{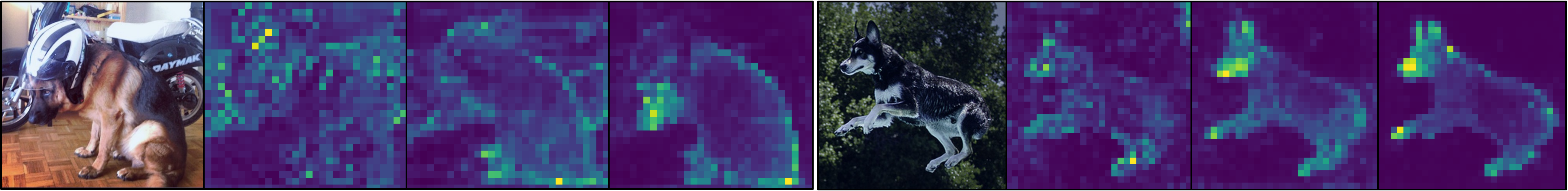}
\vspace{-2.0em}
\caption{Attention visualization of two RoI images produced by MAL. In each image group, the left-most image is the original image. We visualize the attention map output by the $\text{4}^{\text{th}}$, $\text{8}^{\text{th}}$,  $\text{12}^{\text{th}}$ MHSA layers of the Standard ViTs in MAL.}
\label{fig:attention}
\vspace{-0.5em}
\end{figure*}

\begin{table}[]
\centering
\renewcommand{\tabcolsep}{8pt}
\resizebox{\linewidth}{!}{
\begin{tabular}{ccc}
\toprule
 Mask decoder & $(\%)\text{Mask AP}_\text{val}$ & $(\%)\text{Ret.}_\text{val}$ \\ \midrule
Fully connected decoder &  35.5 & 79.2 \\
 Fully convolutional decoder &  36.1 & 80.5 \\
 Attention-based decoder & \textbf{42.3} & \textbf{94.4} \\
 Query-based decoder & - & -\\
 \bottomrule
\end{tabular}
}
\caption{Ablation study of box expansion. We use Standard ViT-MAE-Base as the image encoder of MAL in phase 1 and Cascade RCNN with ConvNext-Small as the instance segmentation models in phase 2. The numbers are reported in \% Mask mAP. Among different designs, the attention-based decoder performs the best. We can not obtain reasonable results with Query-based Decoder.}
\label{tab:mask_decoder}
\vspace{-0.5em}
\end{table}

We compare four different modern designs of mask decoders: the fully connected Decoder~\cite{dai2016instance}, the fully convolutional decoder~\cite{he2017mask,long2015fully}, the attention-based decoder~\cite{bolya2019yolact,wang2020solov2}, and the query-based decoder~\cite{cheng2021mask2former} in Tab. ~\ref{tab:mask_decoder}. We visualize different designs of mask decoders in Figure~\ref{fig:mask_decoder}. For the fully connected Decoder, we use two fully connected layers with a hidden dimension of 2048 and then output a confidence map for each pixel. We reshape this output vector as the 2D confidence map. We introduce the attention-based decoder in Sec~\ref{sec:mal}. For the fully convolutional Decoder, We adopt the pixel-wise head $V$ in the attention-based Decoder. For the query-based decoder, we follow the design in Mask2Former~\cite{cheng2021mask2former}. We spend much effort exploring the query-based Decoder on MAL since it performs extremely well on fully supervised instance segmentation. However, the results are surprisingly unsatisfactory. We suspect the slightly heavier layers might cause optimization issues under the box-supervised losses. 

Experiments show that box-supervised instance segmentation favors the attention-based decoder. However, state-of-the-art instance segmentation and object detection methods often adopt the fully convolutional decoder~\cite{li2022exploring,wang2022image} or the query-based decoder~\cite{cheng2021mask2former}. Our proposed two-phase framework resolves this dilemma and allows the networks to enjoy the merits of both the attention-based Decoder and the non-attention-based Decoders.

\begin{figure*}[t]
\centering
\includegraphics[width=\linewidth]{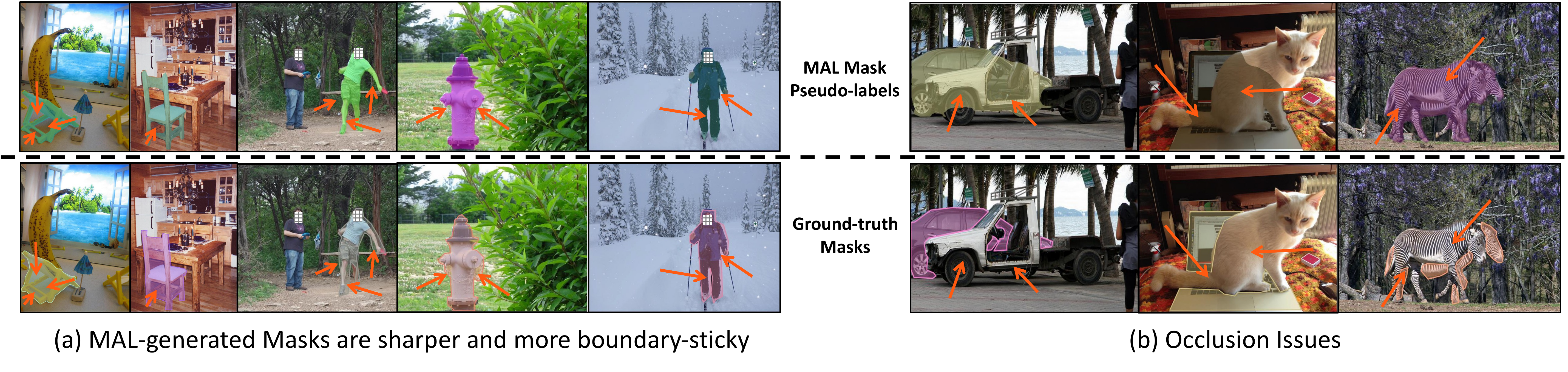}
\vspace{-2.1em}
\caption{The lateral comparison between MAL-generated pseudo-labels (top) and GT masks (bottom) on COCO val2017. On the left, we observe that MAL-generated pseudo-labels are sharper and more boundary-sticky than GT masks in some cases. On the right, we observe that in highly occluded situations, human-annotated masks are still better. }
\label{fig:malvsgt}
\vspace{-1em}
\end{figure*}

\begin{table}[]
\renewcommand{\tabcolsep}{8pt}
\resizebox{\linewidth}{!}{
\centering
\begin{tabular}{lccc}
\toprule
 Backbone  & IN-1k Acc@1 & $\text{Mask AP}_\text{val}$ & $\text{Ret.}_\text{val}$ \\ \midrule
 ConvNeXt-Base~\cite{liu2022convnet}  & \textbf{83.8} & 39.6 & 88.4 \\
 Swin-Base~\cite{liu2021swin} & 83.5 & 40.2 & 89.7 \\
ViT-DeiT-Small~\cite{pmlr-v139-touvron21a} & 79.9 & 40.8 & 91.0  \\
ViT-DeiT-Base~\cite{pmlr-v139-touvron21a} & 81.8 & \textbf{41.1} & \textbf{91.7} \\
\midrule
 ViT-MAE-Base~\cite{he2022masked} & 83.6 & \textbf{42.3} & \textbf{94.4}\\
 ViT-MAE-Large~\cite{he2022masked} & 85.9 & \textbf{42.3} & \textbf{94.4}\\
 \bottomrule
\end{tabular}
}
\caption{Ablation study of different backbones. All models are pre-trained on ImageNet-1k. ConvNeXt and Swin Transformer outperform DeiT on image classification, but standard ViT-Small~\cite{touvron2021training} (ViT-DeiT-Small) outperforms ConvNeXt-base and Swin-Base on mask Auto-labeling. Standard ViT-Base (ViT-MAE-Base) and Standard ViT-Large (ViT-MAE-Large) pretrained via MAE achieve the best performance on mask Auto-labeling.}
\label{tab:encoder_variation}
\vspace{-0.5cm}
\end{table}

\subsection{Clustering analysis}~\label{met:analysis}
\vspace{-1em}

As the results are shown in Tab.~\ref{tab:encoder_variation}, we wonder why the Standard ViTs outperform other modern image encoders in auto-labeling. As the comparison of classification ability does not seem to reflect the actual ability of auto-labeling, we try to use the ability clustering to evaluate the image encoders because foreground(FG)/background(BG) segmentation is very similar to the binary clustering problem. 

Specifically, we extract the feature map output by the last layers of Swin Transformers~\cite{liu2021swin}, ConvNeXts~\cite{liu2022convnet}, Standard ViTs~\cite{dosovitskiy2020image}. Then, we use the GT mask to divide the feature vectors into the FG and BG feature sets. By evaluating the average distance from the FG/BG feature vectors to their clustering centers, we can reveal the ability of the networks to distinguish FG and BG pixels empirically.

Formally, we define the feature vector of token $i$ generated by backbone E as $\bm{f}^E_i$. We define the FG/BG clustering centers $\bm{f}^\prime_1$, $\bm{f}^\prime_0$ as the mean of the FG/BG feature vectors. Then, we use the following metric as the clustering score:

\begin{small}
    \begin{equation}
    S = \frac{1}{N}\sum_i^N (\frac{\bm{f}^E_i}{| \bm{f}^E_i |} - \frac{\bm{f}^\prime_{\gamma(i)}}{| \bm{f}^\prime_{\gamma(i)} |})^2 ,
   \end{equation}
\end{small}

\noindent where if pixel i is FG, $\gamma(i) = 1$, otherwise $\gamma(i) = 0$.

We show the clustering evaluation on the COCO val 2017 in Tab.~\ref{tab:clustering_analysis}. The results align our conclusion that Standard Vision Transformers are better at mask auto-labeling.

\begin{figure}[t]
	\centering
	\includegraphics[width=\linewidth]{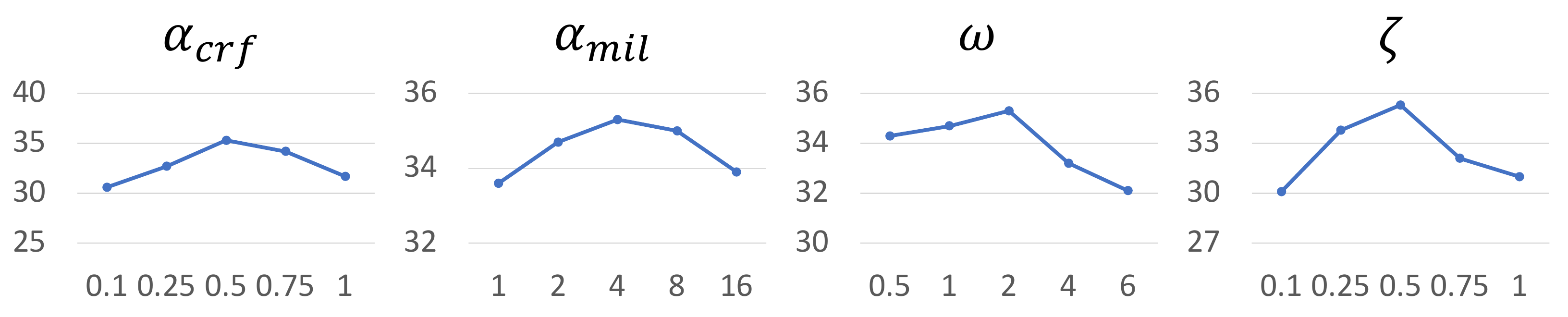}
	\vspace{-1.8em}
	\caption{Sensitivity analysis of loss weights and CRF hyper-parameters. We use ViT-Base~\cite{dosovitskiy2020image} pretrained via MAE~\cite{he2022masked} as the image encoder for the first phase and SOLOv2 (ResNet-50) for the second phase. The x-axis and y-axis indicate the hyper-parameter values and the (\%)mask AP, respectively.}
	\label{fig:sensitivity}
\end{figure}

\begin{table}[!]
\begin{minipage}[t]{0.47\linewidth}
\centering
\resizebox{0.82\linewidth}{!}{
\begin{tabular}{ccc}
\toprule
$\theta$ & $\text{Mask AP}_\text{val}$ & $\text{Ret.}_\text{val}$ \\ \midrule
0.6 &  41.3 & 92.2 \\
 0.8 &  41.7 & 93.1 \\
 1.0 & 42.2 & 94.2 \\
 1.2 & \textbf{42.3} & \textbf{94.4} \\
 1.4 & 42.0 & 93.8 \\
 1.6 & 41.8 & 93.3 \\
 \bottomrule
\end{tabular}
}
\caption{Ablation on box expansion ratio. We use Standard ViT-Base pretrained via MAE (ViT-MAE-Base) and Cascade R-CNN (ConvNeXt-Small) for phase 1 and 2.}
\label{tab:box_expansion}
\end{minipage}
\hspace{4pt}
\begin{minipage}[t]{0.47\linewidth}
\centering
\resizebox{\linewidth}{!}{
\begin{tabular}{lc}
\toprule
 Backbone  & Score ($\downarrow$) \\ \midrule
 ConvNeXt-Base~\cite{liu2022convnet}  & 0.459  \\
 Swin-Base~\cite{liu2021swin} & 0.425  \\
ViT-DeiT-Small~\cite{pmlr-v139-touvron21a} & 0.431  \\
ViT-DeiT-Base~\cite{pmlr-v139-touvron21a} &  0.398  \\
 ViT-MAE-Base~\cite{he2022masked} & 0.324  \\
 ViT-MAE-Large~\cite{he2022masked} & \textbf{0.301}  \\
 \bottomrule
\end{tabular}
}
\caption{Clustering scores for different image encoders. The smaller clustering scores imply a better ability to distinguish foreground and background features.}
\label{tab:clustering_analysis}
\end{minipage}
\vspace{-2em}
\end{table}

\subsection{MAL masks v.s. GT masks}

We show the apples to apples qualitative comparison in Fig.~\ref{fig:malvsgt} and make the following observations. First, MAL-generated mask pseudo-labels are considerably sharper and boundary-sticky than human-annotated ones since humans have difficulties in aligning with the true boundaries. Second, severe occlusion also presents a challenging issue.

\section{Conclusion}

In this work, we propose a novel two-phase framework for box-supervised instance segmentation and a novel Transformer-based architecture, Mask Auto-Labeler (MAL), to generate high-quality mask pseudo-labels in phase 1. We reveal that Standard Vision Transformers are good mask auto-labelers. Moreover, we find that random using box-expansion RoI inputs, the attention-based Decoder, and class-agnostic training are crucial to the strong mask auto-labeling performance. Moreover, thanks to the two-phase framework design and MAL, we can adjust almost all kinds of fully supervised instance segmentation models to box-supervised learning with little performance drop, which shows the great generalization of MAL.

\noindent{\textbf{Limitations.}} Although great improvement has been made by our approaches in mask auto-labeling, we still observe many failure cases in the occlusion situation, where human annotations are much better than MAL-generated masks. Additionally, we meet saturation problems when scaling the model from Standard ViT-Base to Standard ViT-Large. We leave those problems in the future work.

\noindent{\textbf{Broader impacts.}} Our proposed Transformer-based mask auto-labeler and the two-phase architecture serve as a standard paradigm for high-quality box-supervised instance segmentation. If follow-up work can find and fix the issues under our proposed paradigm, there is great potential that expansive human-annotated masks are no longer needed for instance segmentation in the future.

{\small
\bibliographystyle{unsrt}
\bibliography{ref}
}
\appendix
\newpage

\section{Appendix}
\subsection{Additional details of CRF}

In the main paper, we define the energy terms of CRF but skip the details on how we use the Mean Field algorithm to minimize the energy. Here, we provide more details on how we use the Mean Field algorithm~\cite{krahenbuhl2011efficient}.

We define $\bm{l} = \{\bm{l}_1, ..., \bm{l}_N\}$ as the label being inferred, where $N = H \times W$ is the size of the input image and $\bm{x}_i$ is the label of the $i$-th pixel in $\bm{I}$. We also assume that the network predicts a mask $\bm{m} = \{\bm{m}_1, ..., \bm{m}_N\}$ is where $\bm{m}_i$ is the unary mask score of the $i$-th pixel in $\bm{I}$. The pseudo-code to obtain $\bm{l}$ using mean field is attached in Alg.~\ref{alg:1}:

\begin{algorithm}
  \caption{Mean field algorithm for CRFs.}
  \begin{algorithmic}[1]
    \Procedure{MeanField}{$\bm{m}, \bm{I}$}
      \State $\bm{K}_{i,j} \xleftarrow[]{}  \omega \exp(-\frac{|\bm{I}_{i} - \bm{I}_{j}|}{2\zeta^2})$
      \State \Comment{Initialize the Gaussian kernels}
      \State $\bm{l} \gets \bm{m}$ \Comment{Initialize $\bm{l}$ using $\bm{m}$}
      \While{not converge} \Comment{Iterate until convergence}
        \For{$i \gets 1 \,\textbf{to}\, | \bm{l}|$}
            \State $\hat{\bm{l}}_i \gets \bm{l}_i$
            \For{$j \in \mathcal{N}(i)$} 
                \State $\hat{\bm{l}}_{i} \gets \hat{\bm{l}}_{i} + \bm{K}_{j} * \bm{l}_{j}$ 
                \State \Comment{Message passing}
            \EndFor
        \EndFor
        \State $\bm{l} \gets \varphi(\hat{\bm{l}})$ \Comment{$\varphi$ is a clamp function}
      \EndWhile\label{euclidendwhile}
      \State {\textbf{return} $\lambda(\bm{l})$ \Comment{$\lambda$ is a threshold function}}
    \EndProcedure
  \end{algorithmic}
  \label{alg:1}
\end{algorithm}

\subsection{Additional implementation details}
We use the same hyper-parameters on all benchmarks for all image encoders (Standard ViTs~\cite{dosovitskiy2020image, touvron2021training, he2022masked}, Swin Transformers~\cite{liu2021swin}, and ConvNeXts~\cite{liu2022convnet}) and mask decoders (fully connected decoder, fully convolutional decoder, attention-based decoder, ), including batch size, optimization hyper-parameters. We observe a performance drop when we add parametric layers or multi-scale lateral/skip connections~\cite{lin2017feature,li2022exploring} between the image encoder (Standard ViTs, Swin Transformers, ConvNeXts) and the mask decoder (attention-based decoder). We insert a couple of the bi-linear interpolation layers to resize the feature map between the image encoder and the mask decoder and resize the segmentation score map. Specifically, we resize the feature map produced by the image encoder to 1/16 (small), 1/8 (medium), 1/4 (large) size of the raw input according to the size of the objects. We divide the objects into three scales regarding to the area of their bound boxes. We use the area ranges of [0, $32^2$), [$32^2$, $96^2$), [$96^2$, $\infty$) to cover small, medium, and large objects, respectively. We resize the mask prediction map to 512 $\times$ 512 to reach the original resolution of the input images.

Moreover, we also try three naive ways to add classification loss, but it does not work well with MAL. First, we add another fully connected layer as the classification decoder, which takes the feature map of the first fully connected layer of the instance-aware head $K$. With this design, the classification causes a significant performance drop. Secondly, we use two extra fully connected layers or the original classification decoder of standard ViTs as the classification decoder, which directly takes the feature map of the image encoder. However, the classification loss does not provide performance improvement or loss in this scenario.

\subsection{Benefits for object detection}

The supervised object detection models benefit from the extra mask supervision~\cite{he2017mask}, which improves detection results. Specifically, we follow the settings in Mask R-CNN~\cite{he2017mask}. First, we use RoI Align, the box branch, and the box supervision without mask supervision. Second, we add the mask branch and ground-truth mask supervision on top of the first baseline. The second baseline is the original Mask R-CNN. Thirdly, we replace the ground-truth masks with the mask pseudo-labels generated by MAL on top of the second baseline. It turns out that using MAL-generated mask pseudo-labels for mask supervision brings in an improvement similar to ground-truth masks on detection. We show the results in Tab.~\ref{tab:lvis_det_exp}.

\begin{table*}[t]
\centering
\resizebox{\textwidth}{!}{
\addtolength{\tabcolsep}{7pt}
\begin{tabular}{lccccccccc}
\toprule
InstSeg Backbone & Dataset & Mask Labels & $(\%)\text{AP}$ & $(\%)\text{AP}_{50}$ & $(\%)\text{AP}_{75}$ & $(\%)\text{AP}_\text{S}$ & $(\%)\text{AP}_\text{M}$ & $(\%)\text{AP}_\text{L}$ \\
\midrule
ResNet-50-DCN~\cite{he2016deep} & LVIS v1 & None & 22.0 & 36.4 & 22.9 & 16.8 & 29.1 & 33.4  \\
ResNet-50-DCN~\cite{he2016deep} & LVIS v1 & GT mask & 22.5 & 36.9 & 23.8 & 16.8 & 29.7 & 35.0 \\
ResNet-50-DCN~\cite{he2016deep} & LVIS v1 & MAL mask & 22.6 & 37.2 & 23.8 & 17.3 & 29.8 & 34.6  \\
\midrule
ResNet-101-DCN~\cite{he2016deep} & LVIS v1 & None & 24.4 & 39.5 & 26.1 & 17.9 & 32.2 & 36.7 \\
ResNet-101-DCN~\cite{he2016deep} & LVIS v1 & GT mask & 24.6 & 39.7 & 26.1 & 18.3 & 32.1 & 38.3  \\
ResNet-101-DCN~\cite{he2016deep} & LVIS v1 & MAL mask & 25.1 & 40.0 & 26.7 & 18.4 & 32.5 & 37.8 \\
\midrule

ResNeXt-101-32x4d-FPN~\cite{he2016deep,lin2017feature} & LVIS v1 & None & 25.5 & 41.0 & 27.1 & 18.8 & 33.7 & 38.0 \\
ResNeXt-101-32x4d-FPN~\cite{he2016deep,lin2017feature} & LVIS v1 & GT mask & 26.7 & 42.1 & 28.6 & 19.7 & 34.7 & 39.4 \\
ResNeXt-101-32x4d-FPN~\cite{he2016deep,lin2017feature} & LVIS v1 & MAL mask & 26.3 & 41.5 & 28.3 & 19.5 & 34.5 & 39.6 \\

\midrule

ResNeXt-101-64x4d-FPN~\cite{he2016deep,lin2017feature} & LVIS v1 & None & 26.6 & 42.0 & 28.3 & 19.8 & 34.7 & 39.9 \\
ResNeXt-101-64x4d-FPN~\cite{he2016deep,lin2017feature} & LVIS v1 & GT mask & 27.2 & 42.8 & 29.2 & 20.2 & 35.7 & 41.0 \\
ResNeXt-101-64x4d-FPN~\cite{he2016deep,lin2017feature} & LVIS v1 & MAL mask & 27.2 & 42.7 & 29.1 & 19.8 & 35.9 & 40.7 \\

\midrule
ConvNeXt-Small~\cite{liu2022convnet} & COCO & None &  51.5 & 70.6 & 56.1 & 34.8 & 55.2 & 66.9 \\
ConvNeXt-Small~\cite{liu2022convnet} & COCO & GT mask & 51.8 & 70.6 & 56.3 & 34.5 & 55.9 & 66.6 \\
ConvNeXt-Small~\cite{liu2022convnet} & COCO & MAL mask &  51.7 & 70.5 & 56.2 & 35.2 & 55.7 & 66.8 \\
\bottomrule
\end{tabular}
}
\caption{Results of detection by adding different mask supervision. The models are evaluated on COCO val2017 and LVIS v1. By adding mask supervision using ground-truth masks or mask pseudo-labels, we can get around 1\% improvement on different AP metrics on LVIS v1. On COCO val2017, the detection performance also benefits from mask pseudo-labels. Although the improvement is less than COCO's, the improvement is consistent over different random seeds. }
\label{tab:lvis_det_exp}
\end{table*}

\subsection{Additional qualitative results}

We also visualize the prediction results produced by the instance segmentation models trained with ground-truth masks and mask pseudo-labels in Fig.~\ref{fig:mask2former}. In most cases, we argue that humans cannot tell which results are produced by the models supervised by human-annotated labels.

\begin{figure*}[t]
\centering
\includegraphics[width=1.0\linewidth]{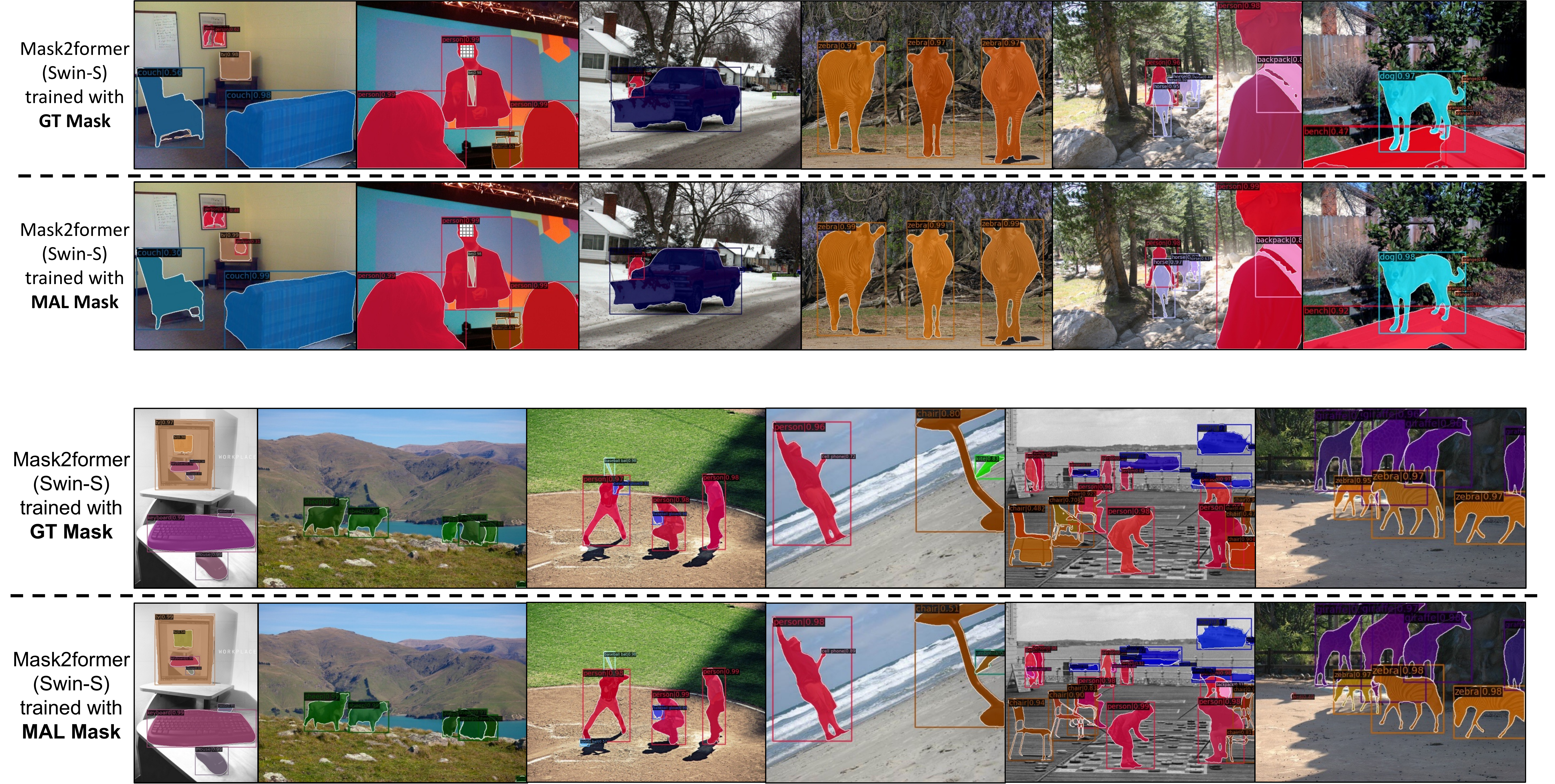}
\caption{The qualitative comparison between Mask2Former trained with GT mask and Mask2Former trained with MAL-generated mask pseudo-labels. Note that we use ViT-MAE-Base as the image encoder of MAL and Swin-Small as the backbone of the Mask2Former.}
\label{fig:mask2former}
\vspace{1.5em}
\end{figure*}

\end{document}